\documentclass[pdf-a,balance,colorlinks,upint,subscriptcorrection,varvw,mathalfa=cal=boondoxo, spanish,french,vietnamese,russian,greek]{asmeconf}

\usepackage{rotating}
\usepackage{mwe}
\usepackage{graphbox}
\usepackage{tabularray}
\usepackage{float}
\usepackage[T1]{fontenc}
\usepackage[utf8]{inputenc}
\usepackage{lmodern}
\usepackage{graphicx}
\usepackage{tikz-cd}

\usepackage{fancyhdr}
\hypersetup{%
	pdfauthor={Leah Chong},									  % <=== change to YOUR name
	pdftitle={IDETC_CAD_GAI},                  % <=== change to YOUR pdf file title
	pdfkeywords={IDETC, CAD, GAI},% <=== change to YOUR pdf keywords
	pdfsubject = {2024 IDETC paper for CAD-guided Generative AI},			  % <=== change to YOUR subject
}

\begin{document}

% \ConfName{Proceedings of the ASME 2024 International Design Engineering Technical Conferences \& \linebreak Computers and Information in Engineering Conference}
% \ConfAcronym{IDETC/CIE 2024}
% \ConfDate{August 25–28, 2024} % update 
% \ConfCity{Washington, DC} % update 
% \PaperNo{DETC2024-146325}

% \title{Navigating the Feasibility-Novelty Tradeoff in Generative Design: Insights from a CAD-Guided Text-to-Image Generation Study}

% \title{A Comparative Analysis of CAD-Guided and Unaided Generative AI in the Context of Engineering Design Feasibility and Novelty}

% \title{The Feasibility-Novelty Nexus in Generative AI Design: Empirical Insights from a CAD-Guided Image Generation Approach}

\title{CAD-Prompted Generative Models: A Pathway to Feasible and Novel Engineering Designs}

% \title{CAD-Guided Text-to-Image Generation and Its Feasibility-Novelty Tradeoff} 

\SetAuthors{%
	Leah Chong\affil{1}\CorrespondingAuthor{lmchong@mit.edu}, 
	Jude Rayan\affil{2}
	Steven Dow\affil{2}, 
	Ioanna Lykourentzou\affil{3},  
	Faez Ahmed\affil{1}
	}

\SetAffiliation{1}{Massachusetts Institute of Technology, Cambridge, MA}
\SetAffiliation{2}{University of California, San Diego, San Diego, CA}
\SetAffiliation{3}{Utrecht University, Utrecht, Netherlands}

%	To switch from inline author names to gridded names, use the [grid] option.

\maketitle


%%%%%  End of fields to be completed. Now write your paper. %%%%%%%%%%%%%%%%%%%%%%%%%%%%%%%%%%%%%%%%%%%

%%%%%  ABSTRACT %%%%%%%%%%%%%%%%%%%%%%%%%%%%%%%%%%%%%%%%%%%%%%%%%%
\begin{abstract}
Text-to-image generative models have increasingly been used to assist designers during concept generation in various creative domains, such as graphic design, user interface design, and fashion design. However, their applications in engineering design remain limited due to the models' challenges in generating images of feasible designs concepts. To address this issue, this paper introduces a method that improves the design feasibility by prompting the generation with feasible CAD images. In this work, the usefulness of this method is investigated through a case study with a bike design task using an off-the-shelf text-to-image model, Stable Diffusion 2.1. A diverse set of bike designs are produced in seven different generation settings with varying CAD image prompting weights, and these designs are evaluated on their perceived feasibility and novelty. Results demonstrate that the CAD image prompting successfully helps text-to-image models like Stable Diffusion 2.1 create visibly more feasible design images. While a general tradeoff is observed between feasibility and novelty, when the prompting weight is kept low around 0.35, the design feasibility is significantly improved while its novelty remains on par with those generated by text prompts alone. The insights from this case study offer some guidelines for selecting the appropriate CAD image prompting weight for different stages of the engineering design process. When utilized effectively, our CAD image prompting method opens doors to a wider range of applications of text-to-image models in engineering design.

\end{abstract}

%%%%%%%%%  BODY OF PAPER %%%%%%%%%%%%%%%%%%%%%%%%%%%%%%%%%

\section{Introduction}
Concept generation is one of the early stages of the engineering design process. During this stage, engineers and designers generate and explore various design concepts \citep{ulrich2019}. Typically followed by the selection of concept(s) to pursue for the rest of the design process, the concepts generated in this stage have been found to have critical impact on the success of the final product. Particularly, the exploration of a wide range of potential concepts through divergent thinking has been encouraged, as it has been found to exercise designers' creativity and enhance the novelty and quality of the final product \citep{cross2021, georgiades2019, hilliges2007}. In design research, various approaches have been proposed to assist designers' divergent thinking, including problem reframing \citep{foster2019}, analogies \citep{mose2017, song2022}, and Ideation Decks \citep{golembewski2010}.

Recently, text-to-image (T2I) generative models, such as DALL-E and Stable Diffusion \citep{ramesh2021, rombach2021, Dayma_DALL·E_Mini_2021}, have shown promise in augmenting designers' concept generation process \citep{brisco2023, lee2023}. They not only enable designers to immediately create visual representations of their ideas but also facilitate the exploration of diverse design concepts. Trained on a large pool of images, T2I models often produce images of design concepts that can inspire designers beyond their own creative capacities \citep{cai2023, Paananen2023, brisco2023}. Therefore, they are increasingly being implemented in various visual art domains \citep{ko2023}, such as graphic design \citep{ueno2021}, user interface design \citep{wang2021, gajjar2021}, fashion design \citep{jeon2021}, and digital art \citep{yurman2022}. 

Despite their widespread reach, the application of T2I models in engineering design remains limited. One reason is that these models frequently generate images of obviously infeasible designs concepts \citep{giannone2023, ada2023, regenwetter2023}. Unlike the visual and textual art contexts, such as graphic design and digital art, engineering design often requires the images of generated concepts to eventually be realized into functional physical products \citep{liu2023}. However, because T2I models are trained on large datasets of texts and images without the understanding of physical constraints, they produce images of creative yet highly infeasible design concepts that cannot be utilized as anything beyond an inspiration \citep{giannone2023, ada2023, regenwetter2023}. For example, T2I models often generate images of unrealistic wheels, such as those that are triangular and/or without any spokes. 

Multiple research efforts have been dedicated to improving the T2I models' suitability for engineering design by enhancing the feasibility of the generated design concepts. One approach is prompt engineering. Some studies have suggested extracting images with three-dimensional (3D) qualities by prompting the T2I models with specific words, such as "3D render" or "CGI" \citep{liu2021, parsons2022}. In addition, Liu, et al. propose 3DALL-E that combines DALL-E, GPT-3, and Contrastive Language–Image Pre-training (CLIP) within a computer aided design (CAD) software called Autodesk Fusion 360 \citep{liu2023}. This method assists designers during their prompt creation process, helping them to streamline the disconnect between the infeasible images and the following stages of the design process involving manufacturing considerations. GPT-3 and CLIP help designers craft multimodal inputs (text and image) that will effectively prompt DALL-E to generate novel 3D images that meet their design goals. Although 3DALL-E encourages designers to think through the manufacturability of a design, it does not directly address the inherent problem of low feasibility of the generated design concepts. Furthermore, this method adds complexity to the generation process, demanding more steps to be taken by the designers to produce feasible design concepts.

Some researchers have also been developing text-to-3D models, such as DreamFusion and Point-E \citep{poole2022, nichol2022, song2023}. These models are specifically meant for engineering design applications to generate feasible 3D models of designs rather than images. Despite their potential, these models currently present many shortcomings. They are much more computationally demanding than the T2I models, requiring longer inference times \citep{jain2021}. They are also restricted in their capability to generate diverse and novel designs due to the smaller volume of accessible training datasets \citep{nichol2022, liu2023, song2023}. These inherent restrictions of text-to-3D models compromise the opportunity to use generative artificial intelligence (AI) to quickly produce novel inspirations \citep{mukherjee2023}. 

In this paper, we propose a CAD image prompting method to improve the feasibility of T2I models' generated design concepts for engineering design applications. Specifically, we work with Stable Diffusion 2.1, exploiting its capability to accept multimodal input, such as texts and images, to create images in a zero-shot manner \citep{ramesh2021}. Given a text prompt, our method first identifies a CAD image of a feasible design. Then, both the text prompt and the identified CAD image are used as prompts for the image generation. To explore the potential of this CAD image prompting method for engineering design, we conducted a case study with a bike design task and investigate the feasibility and novelty of the designs generated by this method.  

%%%%%%%%%%%%%%%%%%%%%%%%%%%%%%%%%%%%%%%%%%%%%%%%%%%%%%%%%%% 

\section{Methods}

\begin{figure}[h!]
\centering{
\includegraphics[height=0.22\textheight, alt = {Linearization error in radiant flux}]{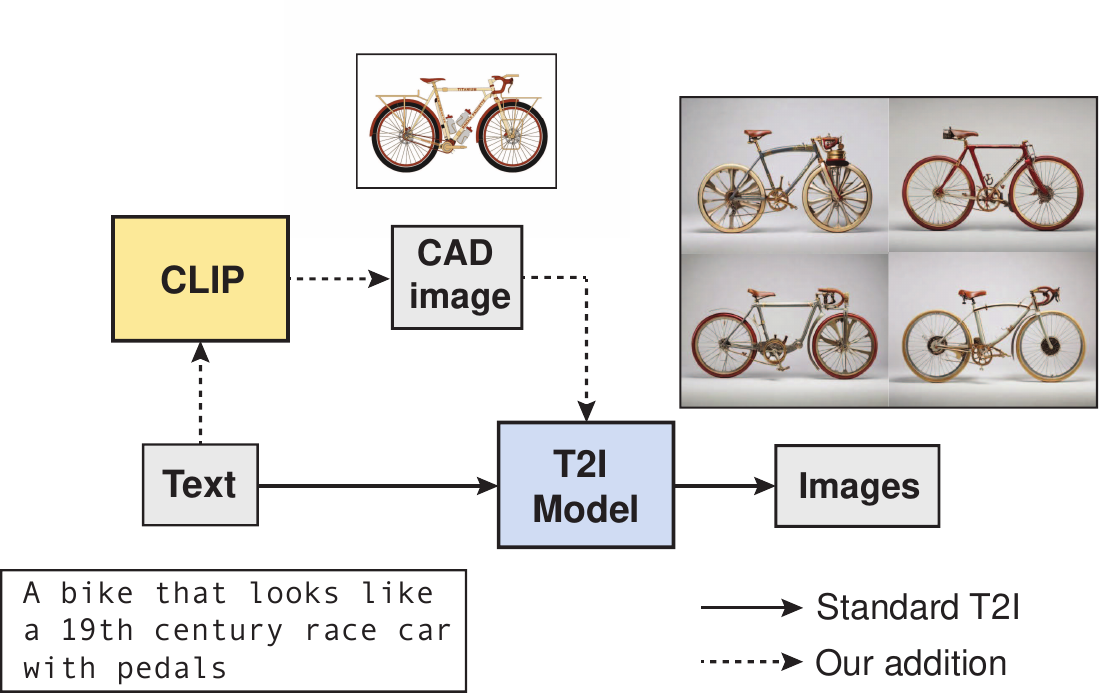}
\caption{Our CAD image prompting method\label{fig:1}. Based on a designer's text prompt, the CLIP model searches for the most suitable CAD image that is then input into a T2I model and guides the image generation process alongside the original text prompt.}
}
\end{figure}

\subsection{CAD Image Prompting for Text-to-Image Generation}

The workflow of our CAD image prompting method is shown in Fig.~\ref{fig:1}. The method relies on a dataset of CAD \textit{images} (not in CAD file formats, such as dwg and dxf) of the product that is being designed. We use images rather than dwg/dxf files to harness the creative advantage of T2I models and image prompting. For example, if the task is to design a chair, the CAD image dataset would consist of a diverse set of chair designs. Then, given a text prompt entered by a designer, OpenAI's Contrastive Language-Image Pre-Training (CLIP) model is utilized to identify the image that is the most semantically similar to the text prompt from the CAD image dataset. The CLIP model is a neural network trained on 400 million image and text pairs to learn visual concepts from text. The selected CAD image is then used as an image prompt, together with the designer's initial text prompt, to guide the T2I model's image generation. The T2I model blends the text and image prompts according to the prompt weights to produce new images. More details about the prompt weights and the generation settings can be found in Section \ref{sec:2.2.1}.

The goal of the CAD image prompting is to generate designs that are more visibly feasible than those that would be generated using text prompts alone. Enabling the generation of more feasible designs can greatly increase the use of T2I models in more physical product design domains like engineering design. While many studies have proposed image guidance and prompting as a way to improve the generated image quality, this method is unique in its use of CAD images to improve the perceived design feasibility. Since the generated output are still images, the purpose of our CAD image prompting is not to achieve the actual feasibility of the designs, but the apparent, perceived feasibility of the design concepts represented by the images. However, this work could serve as a step towards generating actually feasible designs. Furthermore, CAD images offer an advantage of giving designers more flexibility and control to gear the image generation in their desired direction. This is because CAD image datasets can easily be forged using CAD softwares. Synthetic datasets of CAD images can be fabricated strategically to guide the image generation process toward designers' specific needs.

\subsection{Bike Case Study}
We explored the effectiveness of our proposed CAD image prompting method by conducting a case study with a bike design task adopted from Chong, et al.'s study \citep{chong2024}. The task is to create feasible, novel, and aesthetically-pleasing bike design(s) using an off-the-shelf T2I platform called Leonardo.AI \citep{leonardo}, as follows:  

\begin{table*}[h!]
\caption[Table]{The 15 text prompts used for image generation. These prompts are from Chong, et al.'s study \citep{chong2024}.}\label{tab:1}
\centering{%
\begin{tabular}{cp{15cm}}
\toprule
Participant & Prompt \\
\midrule
1 & A bike that looks like a 19th century race car with pedals \\
2 & A bike with crazy design and unique pattern   \\
3 & Electrical bike with low saddle, foldable, backseat, light weight paddle \\
4 & Bike with bottle cage and red chain rings that can carry people \\
5 & Create a city bike with 26 inch wheel with aluminum build up with front light and tail light, saddle no higher than front seat\\
6 & Electrical bike with large comfortable seat and streaming body \\
7 & Please create a rainbow color bike design that is feasible to manufacture and functions as a bike for human (e.g. no triango wheels and with handle), and as unique/novel and aesthetically as possible. There should be at least two wheels, a handle bar, and a seat \\
8 & A bike that resembles a praying mantis. The bike should be green in color and the background should look like a forest for mountain biking. The design should be feasible to manufacture (no triango wheels!) \\
9 & An innovative bike with comfortable saddle, propellers as the wheels and a cover, made with glass, in the future \\
10 & 22nd century bike \\
11 & Aesthetically pleasing tiffany-green racing bike with super thick wheels, LED bulb decoration \\
12 & Fashion bicycle product with sofa chair on the bike saddle  \\
13 & Bike that can fly and dive \\
14 & Create a unique/novel, feasible, aesthetically pleasing, fashionable bike with metal wings, smart screen, mirror and a phone holder  \\
15 & Commute bike for a man with a water bottle holder \\
\bottomrule
\end{tabular}
}
\end{table*}

\begin{quote}
You work for the company 22-century Bike as a bike product designer. Your job is to make one or more new bike designs. Your new bike design(s) should be feasible to manufacture (no triangle wheels!) and as unique/novel and aesthetically pleasing as possible. You can submit multiple designs. 
\end{quote}
This task choice satisfies two main requirements: it concerns the design of an engineering product (i.e., a bike) that eventually needs to be manufactured into a physical, functional object, and it involves both feasibility and novelty as the design goals. With this task, we first generated a diverse set of bike images using Stable Diffusion 2.1 with and without CAD image prompting, then collected human evaluations on the feasibility and novelty of the designs. 

\subsubsection{Image Generation \label{sec:2.2.1}}
To examine how the bike designs generated with and without CAD image prompting differ in their feasibility and novelty, the first step was to generate the images. For this generation, we used the text prompts provided by the 15 participants in Chong, et al.'s human subject study \citep{chong2024}. Because most of these participants entered multiple text prompts, the most comprehensive prompt from each participant (15 total) was selected and revised (e.g., typos and missing words) for our image generation. The 15 prompts are displayed in Table ~\ref{tab:1}.

\begin{figure*}[h!]
\centering\includegraphics[width=0.9\linewidth, alt = {Linearization error in radiant flux}]{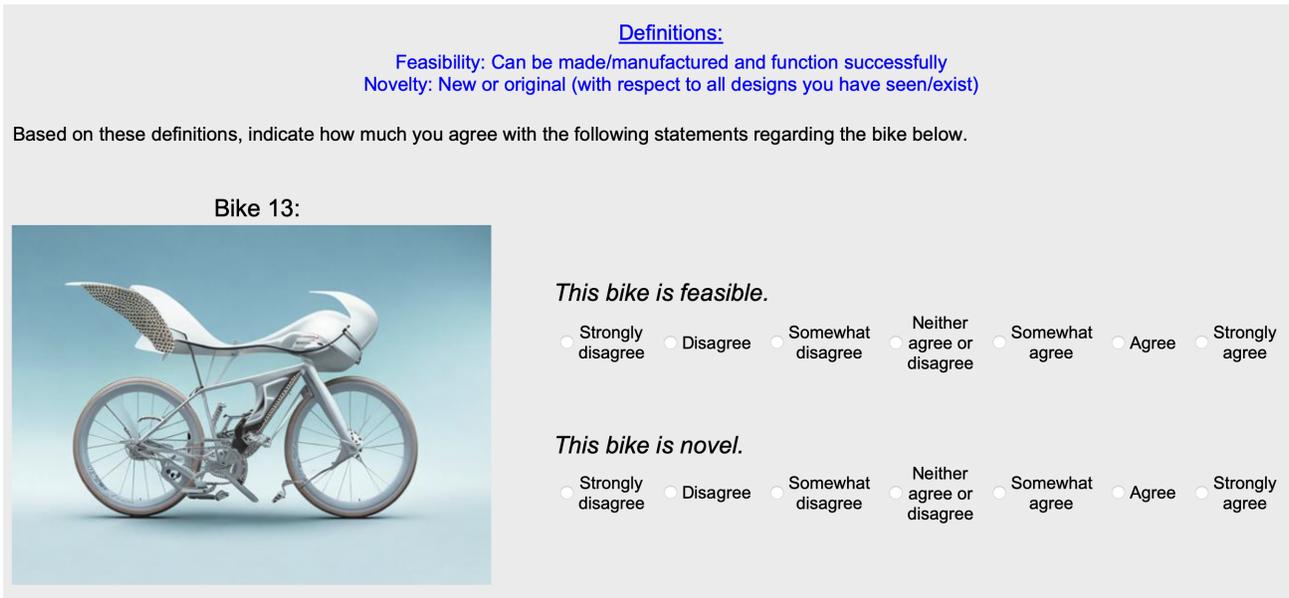}
\caption{Example evaluation questions for a bike design concept\label{fig:2}}
\end{figure*}

Using Leonardo.AI, the authors generated bike design images with Stable Diffusion 2.1 in seven different generation settings:
\begin{quote}
1) Stable Diffusion 2.1 (SD),

2) Stable Diffusion 2.1 and Prompt Magic (SD+PM),

3) Stable Diffusion 2.1, Prompt Magic, and CAD image prompting of weight 0.35 (CIP(0.35)),

4) Stable Diffusion 2.1, Prompt Magic, and CAD image prompting of weight 0.51 (CIP(0.51)),

5) Stable Diffusion 2.1, Prompt Magic, and CAD image prompting of weight 0.67 (CIP(0.67)),

6) Stable Diffusion 2.1, Prompt Magic, and CAD image prompting of weight 0.83 (CIP(0.83)), and

7) Stable Diffusion 2.1, Prompt Magic, and CAD image prompting of weight 1 (CIP(1)). 
\end{quote}
Across the seven settings, the following Leonardo.AI features are kept identical: Number of Images = 4, Input Dimensions = 1024X768, Guidance Scale = 7, Public Images = OFF, PhotoReal = OFF, and Alchemy = OFF. The Number of Images is the quantity of images generated per generation. This value is set to 4 to account for the variation within the same generation. The Input Dimensions refer to the size of the generated images. The Guidance Scale controls how much the image generation process follows the text prompt. We use Leonardo.AI's default value of 7. The Public Images value simply determines whether the generated images will be shared with the public or not. Finally, the PhotoReal and Alchemy options are exclusive features offered by Leonardo.AI that finetune the default T2I model (in our case, Stable Diffusion 2.1) to produce more realistic and high quality images, respectively. To enable our study to be generalizable to any T2I tools other than Leonardo.AI, we have these features off for all image generations. For more information about the features on Leonardo.AI, please refer to their website \citep{leonardo}.

Prompt Magic is an exclusive feature on Leonardo.AI that enhances prompt conformity and image structure. Although this feature is uniquely offered through Leonardo.AI, we have it on because the goal of the feature is to improve the feasibility of images, aligning with the goal of our method. Comparing the feasibility of the design concepts generated in the SD+PM setting without the CAD image prompting to those in the other settings with the CAD image prompting can show whether the CAD image prompting is effective in addition to Prompt Magic or not. When Prompt Magic is on, its High Contrast mode is automatically turned on with the default Prompt Strength of 0.3, which are maintained in our generations. 

In the CIP settings with varying prompting weights, the CAD image is searched from Regenwetter, Curry, and Ahmed's BIKED dataset of 4800 diverse bike images \citep{regenwetter2021}. The selected image is the one that semantically best matches the text prompt. Examples of these selected CAD images are shown in Table ~\ref{tab:2}. The possible range of the CAD image prompting weight offered on the Leonardo.AI platform are from 0.35 to 1. To investigate how the feasibility and novelty of generated images vary with the image prompting weight, the five weight settings (CIP(0.35), CIP(0.51), CIP(0.67), CIP(0.83), and CIP(1)) were determined by uniformly distributing the strength values within the possible range.

\begin{table*}[h!]
\caption{Example images from the seven generation settings (refer to Section \ref{sec:2.2.1})}\label{tab:2}%
% \bigskip
\centering{
\begin{tabular}
{m{0.12\textwidth}|m{0.22\textwidth}|m{0.22\textwidth}|m{0.22\textwidth}}
\toprule
Text prompt \centering & An innovative bike with comfortable saddle, propellers as the wheels and a cover, made with glass, in the future & Fashion bicycle product with sofa chair on the bike saddle & Create a unique/novel, feasible, aesthetically pleasing, fashionable bike with metal wings, smart screen, mirror and a phone holder \\
\midrule
SD\centering & \includegraphics[height=0.1\textheight]{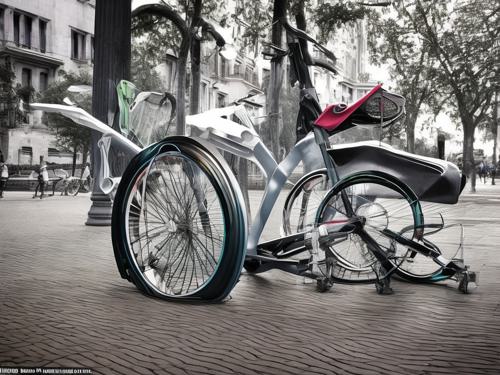}& \includegraphics[height=0.1\textheight]{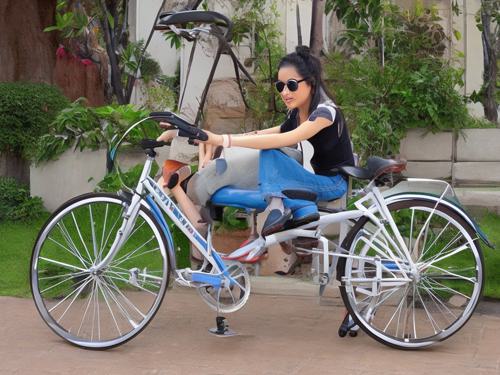} & \includegraphics[height=0.1\textheight]{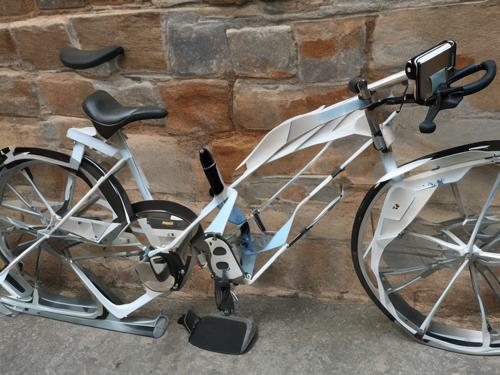} \\
SD+PM\centering & \includegraphics[height=0.1\textheight]{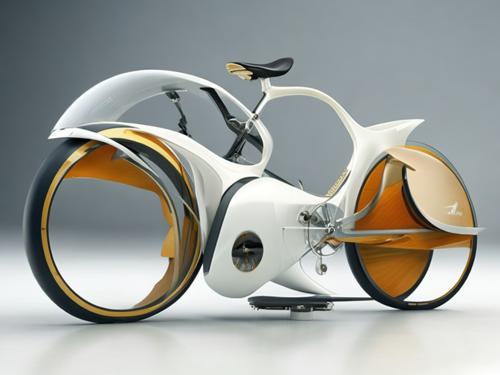} & \includegraphics[height=0.1\textheight]{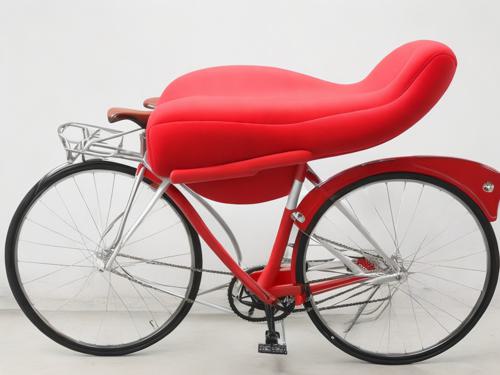} & \includegraphics[height=0.1\textheight]{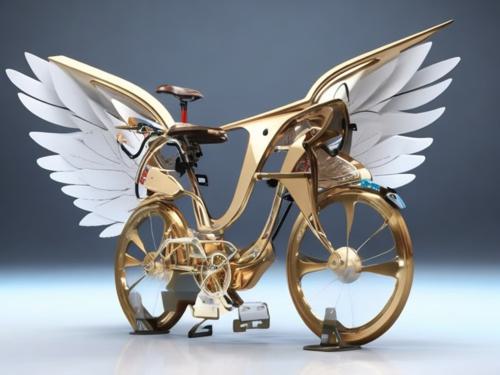} \\ \hline
CAD image prompt\centering & \includegraphics[height=0.08\textheight]{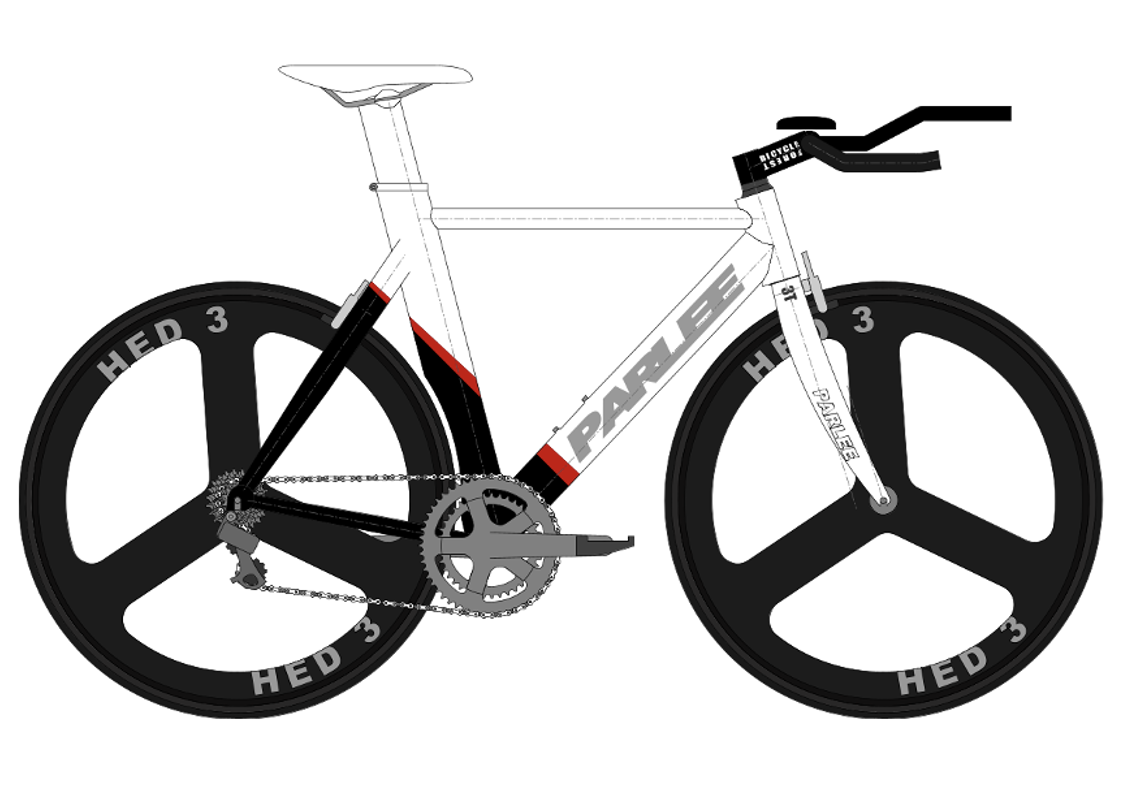} & \includegraphics[height=0.08\textheight]{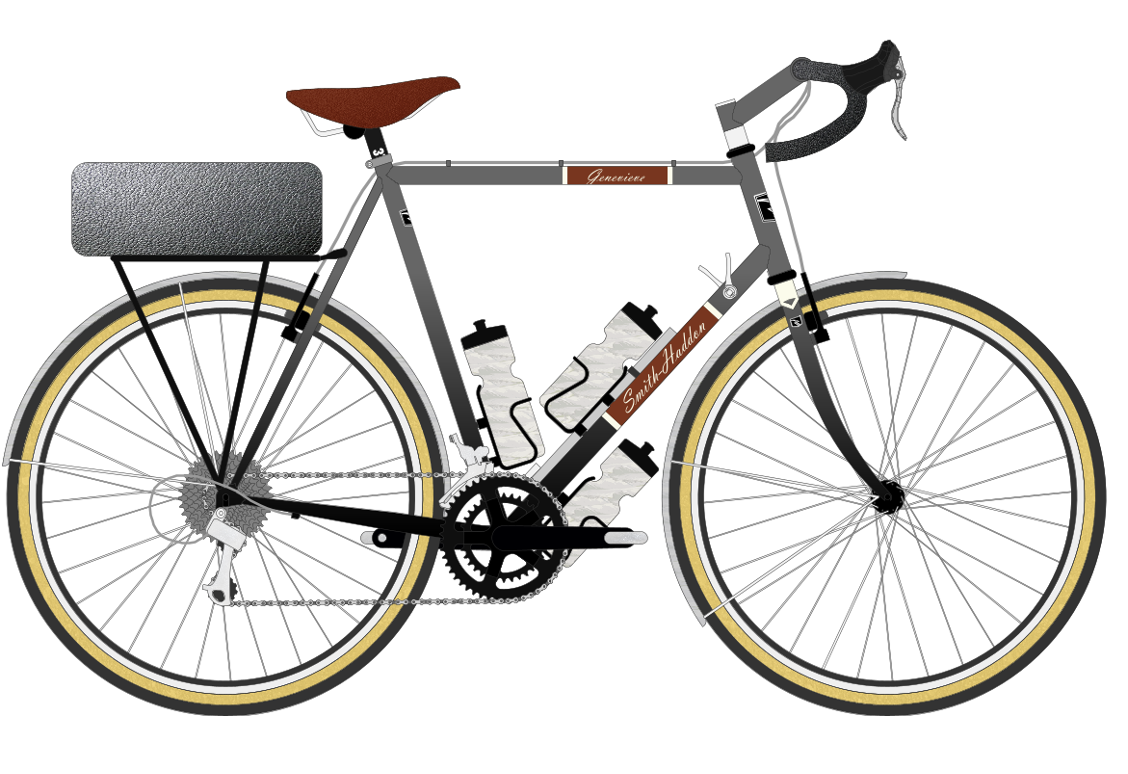} & \includegraphics[height=0.08\textheight]{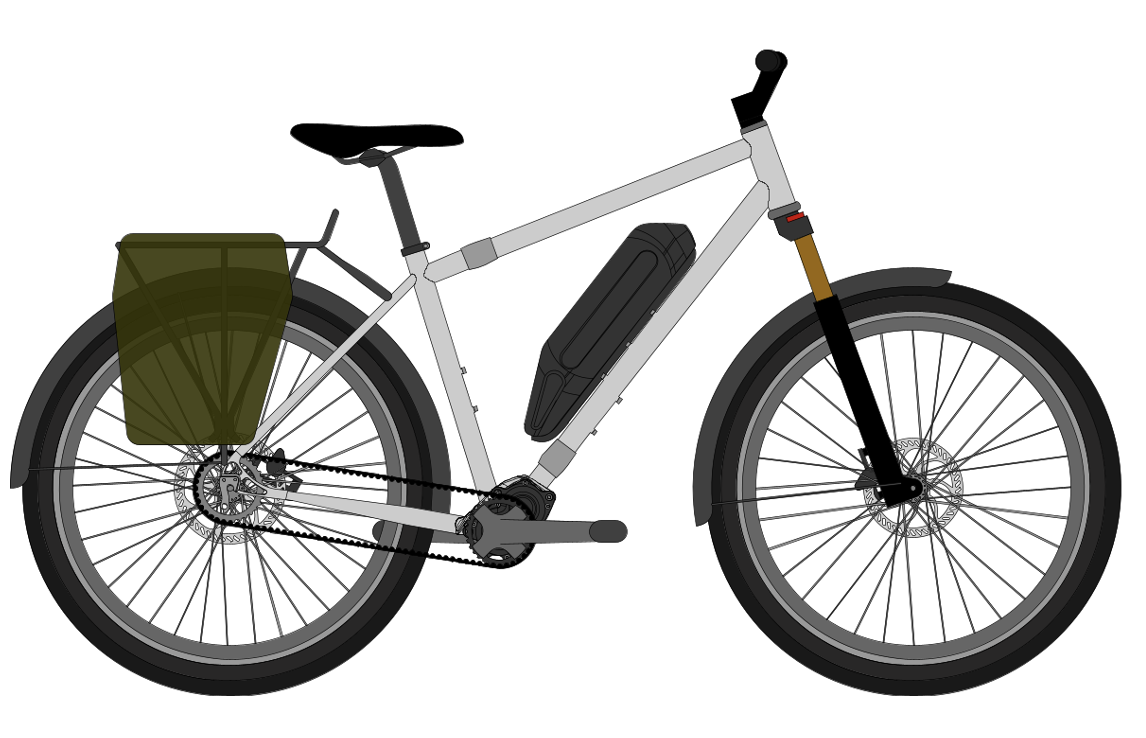} \\ \hline
CIP(0.35)\centering & \includegraphics[height=0.1\textheight]{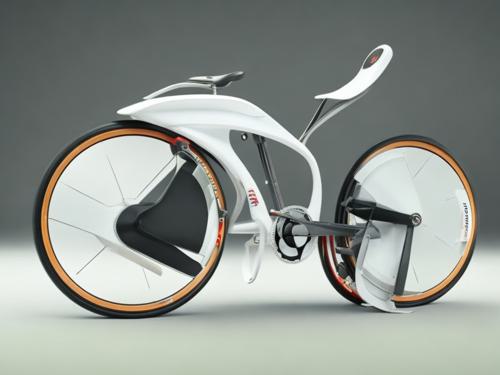} & \includegraphics[height=0.1\textheight]{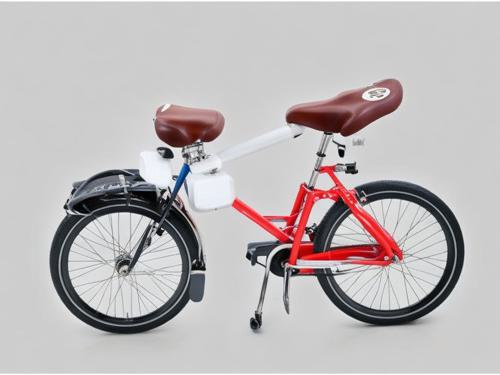} & \includegraphics[height=0.1\textheight]{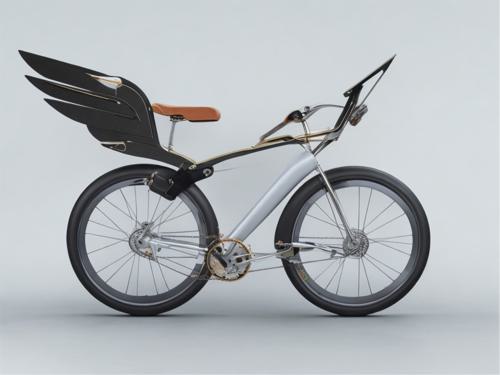} \\
CIP(0.51)\centering & \includegraphics[height=0.1\textheight]{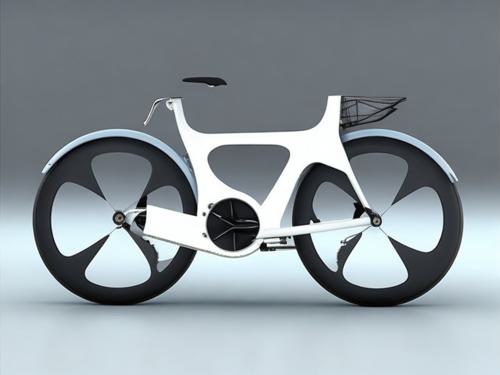} & \includegraphics[height=0.1\textheight]{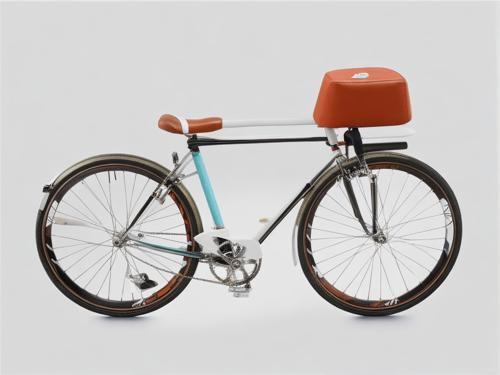} & \includegraphics[height=0.1\textheight]{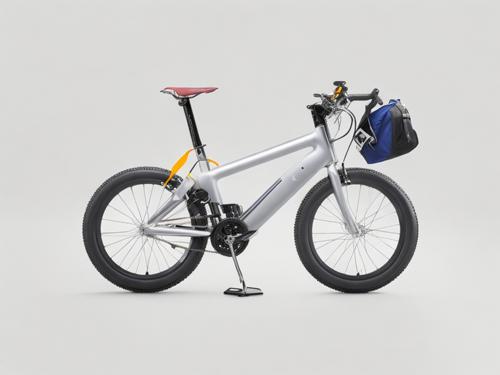} \\
CIP(0.67)\centering & \includegraphics[height=0.1\textheight]{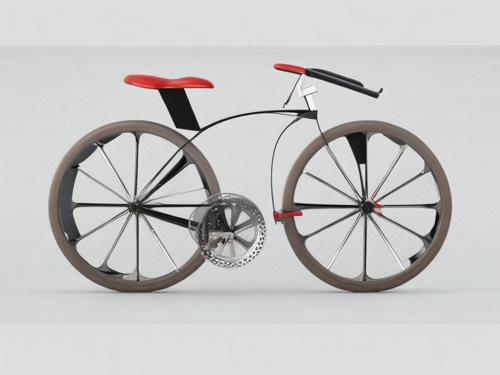} & \includegraphics[height=0.1\textheight]{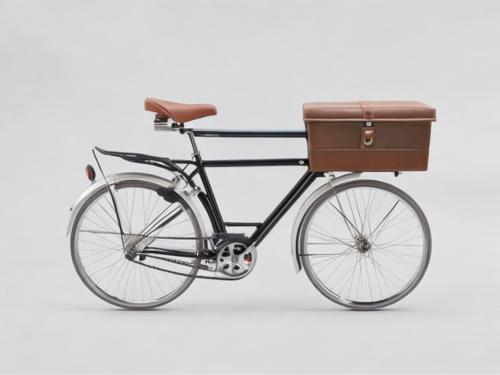} & \includegraphics[height=0.1\textheight]{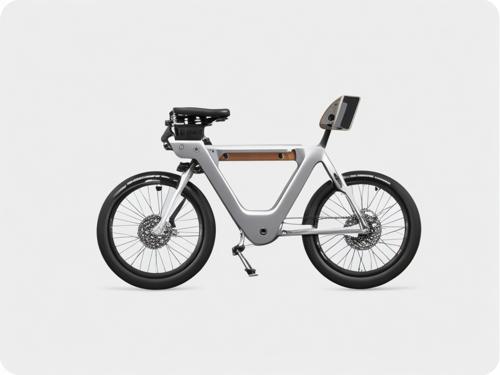} \\
CIP(0.83)\centering & \includegraphics[height=0.1\textheight]{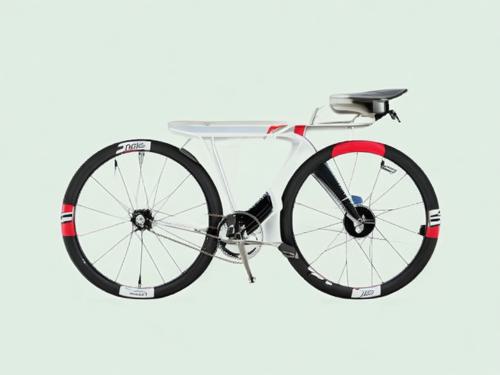} & \includegraphics[height=0.1\textheight]{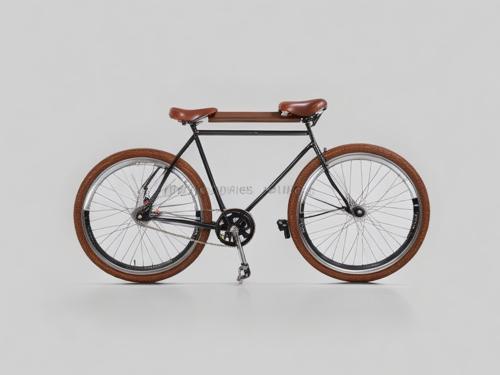} & \includegraphics[height=0.1\textheight]{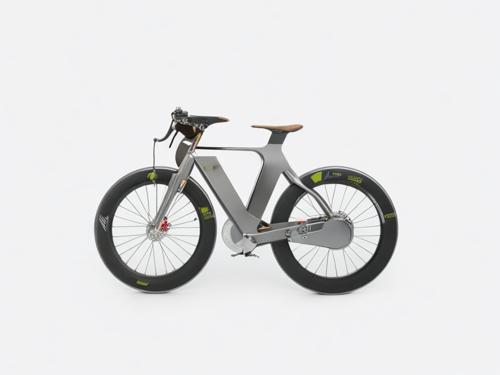} \\
CIP(1)\centering & \includegraphics[height=0.1\textheight]{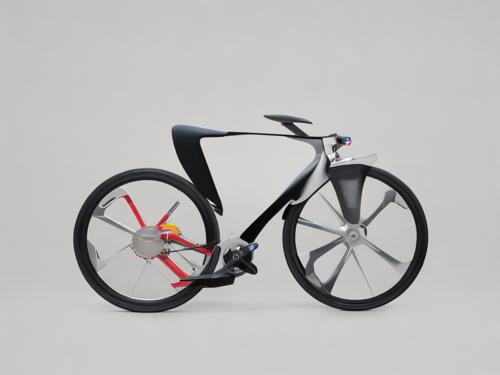} & \includegraphics[height=0.1\textheight]{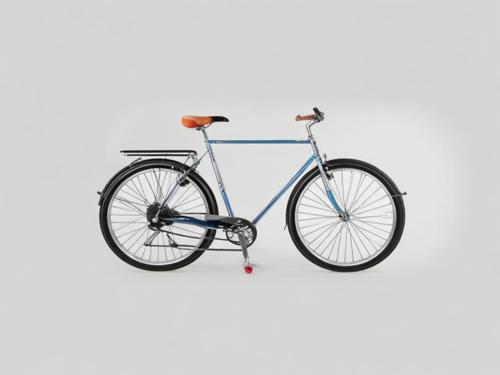} & \includegraphics[height=0.1\textheight]{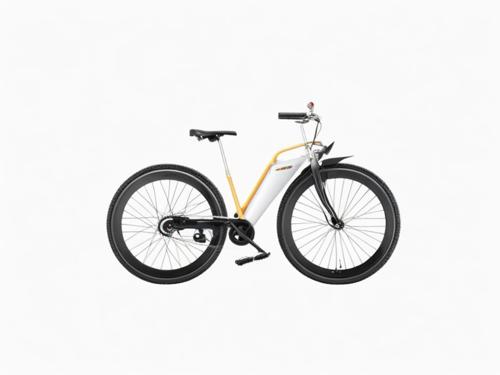} \\
[2pt]
\bottomrule\end{tabular}
}
\end{table*}

In total, 420 images of different bike designs were generated by the authors. For each of the 15 text prompts, 28 images were produced (7 settings per text prompt X 4 images per setting). Table ~\ref{tab:2} illustrates some example images generated in the seven settings, showing one of the four images generated per text prompt and setting. It is important to note that the images generated with the CAD image prompting weight of 1 do not exactly resemble the CAD image. This is because the T2I model is simultaneously trying to adhere to the text prompt, resulting in some variations from the CAD image prompt.

\subsubsection{Design Evaluations}
Having had generated the 420 images of bike designs with 15 text prompts and seven different settings, we collected human evaluations of the perceived feasibility and novelty of each design concept. Since the generated output are still images in this work, the purpose of our CAD image prompting is not to achieve the actual feasibility of the designs, but the apparent, perceived feasibility of the design concepts represented by the images. However, this work could serve as a step towards generating actually feasible designs. The evaluation questions for each design are illustrated in Fig.~\ref{fig:2}. The definitions of feasibility and novelty were provided at the top of the screen for reference so that all participants have the same understanding of the evaluation categories. The definitions were taken and modified from the Oxford Languages. Participants were asked to indicate their degree of agreement or disagreement with the two following statements: 1) "The bike is feasible." and, 2) "The bike is novel", in a 7-point Likert scale from "Strongly disagree" to "Strongly agree". 

After a round of pilot study, it was determined that a participant could evaluate about 28 images in 5 minutes, leading to a decision to give each participant 140 bike designs to evaluate during a 30-minute session. 32 participants were recruited and completed the evaluations, and the data from two of these participants were discarded and replaced because they failed to follow the instructions and/or completed the evaluations too quickly. The participants were recruited using the convenience, voluntary, and snowball sampling methods with the only recruitment criterion being they are 18 or order. We did not require the participants to have any prior design experience (although some did) because the goal was to collect common users' ratings on their perceived feasibility and novelty, rather than objective ratings. Each participant received 140 images randomly selected from 420 images and evaluated them in a random order. Each image was assigned to 10 different participants to be able to obtain the average feasibility and novelty ratings per image.  
% \begin{table*}[h!]
% \caption{The CLIP similarity score matrix between the seven generation settings using Text Prompt 1 in Table \ref{tab:2}. The similarity scores range from -1 to 1, -1 meaning maximum dissimilarity between the images generated in those settings and 1 meaning maximum similarity. The results show that the CAD image prompting significantly affects the generated images.}\label{tab:3}%
% % \bigskip
% \centering{
% \begin{tabular}
% {c|ccccccc}
% \toprule
%  & SD & SD+PM & CIP(0.35) & CIP(0.51) & CIP(0.67) & CIP(0.83) & CIP(1)\\
% \midrule
% SD& 0.370	&-0.235	&-0.215	&-0.496	&-0.0871	&-0.213	&-0.242\\
% SD+PM&	-0.235&0.744	&0.475	&0.308	&0.288	&0.446	&-0.152\\
% CIP(0.35)& -0.215	&0.475	&0.638	&0.475	&0.413	&0.540	&0.0754\\
% CIP(0.51)&-0.496	&0.308	&0.475	&0.660	&0.274	&0.604	&-0.0851\\
% CIP(0.67)&-0.0871	&0.288	&0.413	&0.274	&0.593	&0.470	&0.382\\
% CIP(0.83)&-0.213	&0.446	&0.540	&0.604	&0.494	&0.801	&0.249\\
% CIP(1)&-0.242	&-0.152	&0.0754	&-0.0851	&0.382	&0.249	&0.741\\
% \bottomrule
% \end{tabular}
% }
% \end{table*}
\begin{table*}[h!]
\caption{The CLIP similarity score matrix between the seven generation settings using Text Prompt 1 in Table \ref{tab:2}. The similarity scores range from -1 to 1, -1 meaning maximum dissimilarity between the images generated in those settings and 1 meaning maximum similarity. The results show that the CAD image prompting significantly affects the generated images.}\label{tab:3}%
% \bigskip
\centering{
\begin{tabular}
{c|ccccccc}
\toprule
 & SD & SD+PM & CIP(0.35) & CIP(0.51) & CIP(0.67) & CIP(0.83) & CIP(1)\\
\midrule
SD& 0.370	&-0.235	&-0.215	&-0.496	&-0.0871	&-0.213	&-0.242\\
SD+PM&	&0.744	&0.475	&0.308	&0.288	&0.446	&-0.152\\
CIP(0.35)& 	&	&0.638	&0.475	&0.413	&0.540	&0.0754\\
CIP(0.51)&	&	&	&0.660	&0.274	&0.604	&-0.0851\\
CIP(0.67)&	&	&	&	&0.593	&0.470	&0.382\\
CIP(0.83)&	&	&	&	&	&0.801	&0.249\\
CIP(1)&	&	&	&	&	&	&0.741\\
\bottomrule
\end{tabular}
}
\end{table*}

\section{Results}
In this section, the CLIP similarity scores between the images generated in the seven generation settings (listed in Section \ref{sec:2.2.1}) are presented first in Table \ref{tab:3}. Each CLIP similarity score represents how similar the images generated in the column setting are to those generated in the row setting. The CLIP similarity score between the images generated in two settings is calculated by averaging the similarity scores for all pairs of images. For example, given a prompt, each setting generated four images. For the two sets of four images produced by two settings, the CLIP similarity scores are calculated for every pairwise combination and are averaged. The scores range from -1 to 1, -1 meaning maximum dissimilarity and 1 meaning maximum similarity. Notice that the left diagonal values are not 1, even though they are comparing between the same settings. This is because there are multiple images generated in each setting, and the average of the similarity scores between these images are reported. The lowest similarity scores (mostly negative) are mainly observed for the SD setting, confirming that the images generated without the CAD image prompting are highly dissimilar to the rest of the images. Table \ref{tab:3} displays the results only for the generations with Text Prompt 1 from Table \ref{tab:2}. However, similar trends are observed with the other 14 text prompts.

We also present the human evaluation results for how the perceived feasibility and novelty of the generated bike designs vary among the seven generation settings. These results bring insight into the effectiveness the CAD image prompting method in producing more visibly feasible designs compared to the T2I models with only text prompts, as well as whether and how the novelty of the designs may be compromised. Figure \ref{fig:3} shows the overall results from the human evaluations and the Spearman's rho correlation result between the weight of the CAD image prompt and the feasibility and novelty ratings of the generated bike designs. Table \ref{tab:4} consists of the Mann-Whitney U test results to show whether the feasibility and novelty ratings of the generated designs are different between various pairs of settings.

\subsection{Perceived feasibility increases with CAD image prompting}
\begin{figure*}[h!]
\centering{
\begin{subfigure}[b]{\columnwidth}% subfigure is basically the same as minipage
\centering{
  \includegraphics[width=0.9\linewidth, alt={Nusselt number data for isothermal wall}]{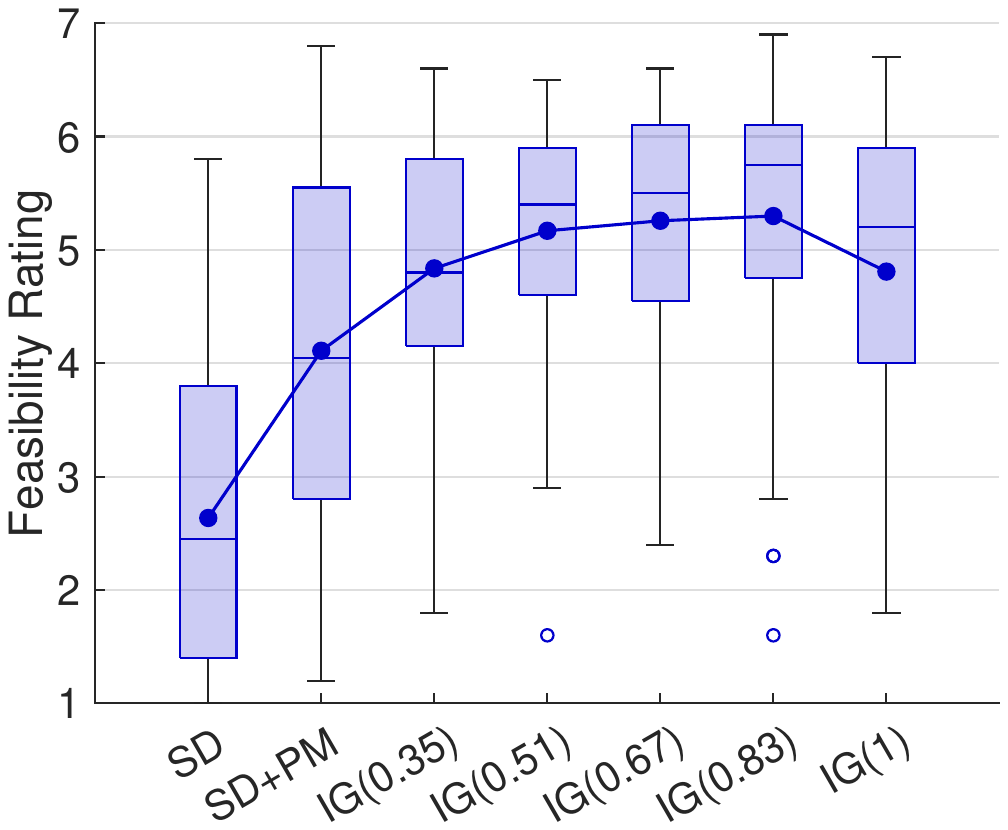}%
}
\subcaption{Perceived feasibility\label{fig:3a}}
\end{subfigure}%
% \hspace*{\columnsep}% with this space added, puts each figure at column center
\begin{subfigure}[b]{\columnwidth}
\centering{%
\includegraphics[width=0.9\linewidth, alt={Nusselt number data for constant heat flux wall}]{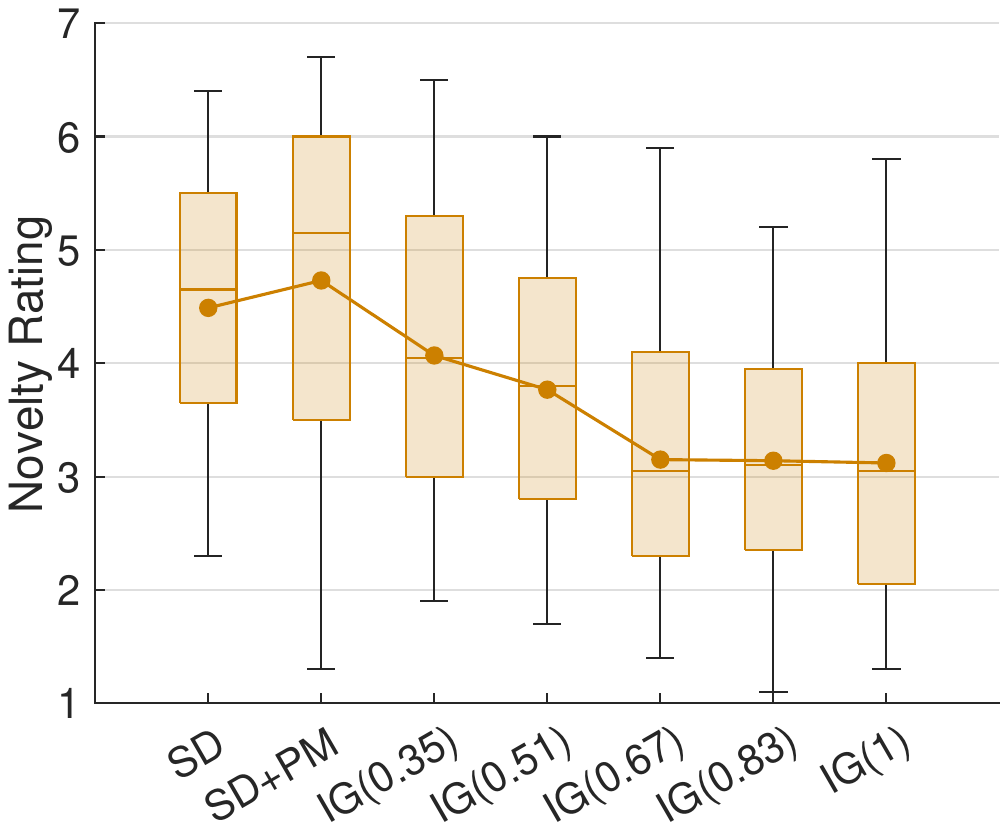}%
}%
\subcaption{Novelty\label{fig:3b}}
\end{subfigure}
\caption{Perceived feasibility and novelty evaluation results for the bike designs generated in the seven generation settings. (a) shows a positive correlation between the weight of CAD image prompt and the perceived feasibility of generated images, while (b) shows a negative correlation. \label{fig:3}}
}
\end{figure*}

Figure \ref{fig:3a} and the first two columns of Table \ref{tab:4} display results related to the feasibility of the generated images. First and foremost, the second and seventh rows in Table \ref{tab:4} demonstrate the effectiveness of our CAD image prompting method in improving the perceived feasibility of generated designs. The feasibility rating is significantly higher in the CAD image prompting settings with 0.35 weight (CIP(0.35)) than Stable Diffusion 2.1 (SD) and Stable Diffusion 2.1 with Prompt Magic (SD+PM) settings (Z = -7.22 and -2.81, p 
% = 5.21E-13 and 4.93E-3, 
< 0.01 and < 0.01, respectively). This result holds true with all the other CAD image prompting settings with different weights, as demonstrated in rows three to six and rows eight to 11 in Table \ref{tab:4}. 

For the Spearman's rho test, the SD+PM setting, rather than the SD setting, is used as the no CAD image prompting (weight = 0) setting. This is because the images generated in the SD+PM setting tend to be perceived as more feasible than those generated in the SD setting (Z = -5.09, p 
% = 3.62E-7
< 0.01). Overall, a positive correlation is found between the weight of the CAD image prompt and the perceived feasibility of the generated images of bike designs ($\rho$ = 0.186, p 
% = 3.88E-4
< 0.01). This means that as we increase the weight of the CAD image prompt, the generated bike designs become more visibly feasible. However, it is important to note that the feasibility rating of the designs drop from the weight 0.83 to 1. This suggests that at a certain weight above 0.83, the T2I model’s ability to balance adherence to the CAD image with the text prompt may be compromised.

\begin{table*}[h!]
\caption{Difference in the feasibility and novelty ratings between various pairs of generation settings}\label{tab:4}%
\centering{%
\begin{tabular*}{0.8\textwidth}{@{\hspace*{1.5em}}@{\extracolsep{\fill}}c!{\hspace*{3.em}}cc!{\hspace*{3.em}}cc@{\hspace*{1.5em}}}
\toprule
\multicolumn{1}{@{\hspace*{4.5em}}l}{Settings} &
\multicolumn{1}{c}{Z (Feasibility)} &
\multicolumn{1}{c!{\hspace*{3.em}}}{p (Feasibility)} &
\multicolumn{1}{c}{Z (Novelty)} &
\multicolumn{1}{c@{\hspace*{1.5em}}}{p (Novelty)} \\ 
\midrule
SD : SD+PM &-5.09&	<0.01&	-1.38&	0.168 \rule{0pt}{11pt}\\
SD : CIP(0.35) &-7.22&	<0.01&	1.72&	0.0859 \rule{0pt}{11pt}\\
SD : CIP(0.51) &-7.89	&<0.01	&3.32 &	<0.01\rule{0pt}{11pt} \\
SD : CIP(0.67) &-7.96	&<0.01	&5.59	&<0.01\rule{0pt}{11pt}\\
SD : CIP(0.83) &7.84	&<0.01	&5.82	&<0.01\rule{0pt}{11pt}\\
SD : CIP(1) &-6.89	&<0.01	&5.66	&<0.01\rule{0pt}{11pt}\\
SD+PM : CIP(0.35) &-2.81&	<0.01&	2.62&	<0.01 \rule{0pt}{11pt}\\
SD+PM : CIP(0.51)&-3.87&	<0.01	&3.80	&<0.01\rule{0pt}{11pt}\\
SD+PM : CIP(0.67)&-4.16	&<0.01	&5.56	&<0.01\rule{0pt}{11pt}\\
SD+PM : CIP(0.83)&-4.27	&<0.01	&5.64	&<0.01\rule{0pt}{11pt}\\
SD+PM : CIP(1)&-2.54	&<0.05	&5.61	&<0.01\rule{0pt}{11pt}\\
CIP(0.35) : CIP(0.51) &-1.56&	0.119&	1.22&	0.222 \rule{0pt}{11pt}\\
CIP(0.51) : CIP(0.67) &-0.696&	0.486&	2.78&	<0.01 \rule{0pt}{11pt}\\
CIP(0.67) : CIP(0.83) &-0.686&	0.493&	-0.0630&	0.950 \rule{0pt}{11pt}\\
CIP(0.83) : CIP(1) &2.31&	<0.05&	0.255&	0.799 \rule{0pt}{11pt}\\
[2pt]
\bottomrule\end{tabular*}
}
\end{table*}

\subsection{Novelty is compromised with higher CAD image prompting weights}
Having confirmed that our proposed method successfully improves the perceived feasibility of the generated designs, we heed to how their novelty is affected with the CAD image prompting. The novelty related results are illustrated in Fig. \ref{fig:3b} and the last two columns of Table \ref{tab:4}. 

With the implementation of CAD image prompting with the lowest weight (0.35), the novelty of the generated bike designs is lower compared to the SD+PM setting but not the SD setting (Z = 2.62 and 1.72, p < 0.01 and = 8.59E-2
% = 8.79E-3 and 0.0859
, respectively). This result shows that the novelty performance of Stable Diffusion 2.1 is not significantly affected by the CAD image prompt of weight 0.35. However, as shown in Fig. \ref{fig:3b}, when Stable Diffusion 2.1 is augmented by the Prompt Magic feature in Leonardo.AI, the novelty of the generated designs slightly increases (though not statistically significant). Therefore, the SD+PM setting produces more novel designs than the CIP(0.35) setting. 

With CAD image prompt of a weight greater than 0.35 (more accurately a weight greater than a value between 0.35 and 0.51), the novelty of the generated designs is compromised. This result is bolstered by rows three to six and rows eight to 11 in Table \ref{tab:4}. The novelty ratings in settings CIP(0.51), CIP(0.67), CIP(0.83), and CIP(1) are all significantly lower than those in settings SD and SD+PM.

Like the feasibility analysis, the SD+PM setting, rather than the SD setting, is used as the no CAD image prompting setting (CIP(0)) for the Spearman's rho test. A negative correlation is observed between the weight of CAD image prompt and the novelty of the generated images of bike designs ($\rho$ = -0.386, p 
% = 3.04E-14
< 0.01). This means that as we increase the weight of the CAD image prompt, the generated bike designs become less novel. Therefore, recalling the earlier result that the generated designs are perceived to be more feasible with a higher CAD image prompting weight, we find that there is a significant tradeoff between the feasibility and novelty ($\rho$ = -0.728, p 
% = 1.16E-60
< 0.01).

\section{Discussion}
In engineering design, T2I models have mainly been used as an inspirational tool during concept generation \cite{lee2023,cai2023,Paananen2023}. However, they are difficult to implement for any purpose beyond inspiration because the generated images of design concepts are often infeasible \citep{giannone2023, ada2023, regenwetter2023}. Not only does the perceived infeasibility of the designs prevent generation of realistic concepts, but also hinder the potential use of the T2I models during the other stages of the engineering design process. To address this issue, we propose in this paper a CAD image prompting method in which CAD images of feasible designs are used as prompts in the image generation process along with text prompts. 

A case study with a bike design task was conducted to explore the effectiveness of this method. Results from the study show that CAD image prompting successfully improves the perceived feasibility of designs generated by Stable Diffusion 2.1 with text prompts alone. The perceived feasibility is also improved compared to the advanced version of Stable Diffusion 2.1 which had been further trained by Leonardo.AI's Prompt Magic feature. This result implies that our proposed method has the potential to expand the applicability of T2I models, so that they can play a role in engineering design by creating more visibly feasible design concepts that are beyond mere inspiration. For instance, while designers currently need to independently produce models for their design concepts after being inspired by the images generated by T2I models, this process can be significantly streamlined when the generated designs are more feasible. The designs represented in the generated images will more likely be able to be realized directly into a physical product, meaning that the images will act as models. Furthermore, the improved feasibility of generated images may increase the usefulness of T2I models in later stages of the engineering design process. Our method may be able to generate images of overall designs or even designs of specific parts feasible enough to be directly used for the system-level design, detailed design, and refinement stages. Although this work is a proof-of-concept of the CAD image prompting method and has only improved the \textit{perceived} feasibility of the generated design concepts, future refinement of this method has the potential to further increase the perceived feasibility and even increase the actual feasibility of the designs. 

Our image prompting method utilizes CAD images rather than real-world images, offering unique advantages in engineering design applications. First, it will be highly synergistic with image-to-CAD or image-to-3D methods that aim to instantly transform images into CAD/3D models \citep{alpha3D, 3DFY, poole2022, nichol2022, song2023}. The images of design concepts generated by our method will likely have characteristics similar to those of CAD images, and therefore, they may more easily be transformed into CAD models. Merging image-to-CAD methods with our CAD image prompting method will engender a design tool that allows designers to shift between concept generation and design refinement anytime. In the image mode, the tool can harness the creative potential from T2I models' large dataset, while in the CAD mode, the designers will be able to make detailed changes to the design. Additionally, CAD image datasets can easily be fabricated using CAD software, giving designers more flexibility and control over the image generation process. When using real-world images, designers need to communicate their design ideas through text if the ideas do not already exist in the world. With our method, these design ideas can be communicated using CAD images. Not only does this make our method extremely versatile, but it also enables better control over the dataset and image generation. 

While the potential of the CAD image prompting in engineering design is clear, our results also highlight the tradeoff between the perceived feasibility and novelty, confirming the difficulty in creating both feasible and novel designs with generative AI \citep{mukherjee2023, regenwetter2022}. The generated designs tend to become less novel with higher CAD image prompt weight. Hence, it is critical to select the appropriate image prompting setting for a given design stage and scenario. Generally, designers engage in an iterative process while working with T2I models to figure out the right balance of settings in order to attain a reliable final design \citep{oppenlaender2022creativity}. The findings in this work regarding the CAD image prompting method drastically reduce the need for multiple iterations to generate feasible bike designs. For example, if a designer is using a T2I model for inspiration during concept generation, it is important for the images to be novel rather than feasible. In this case, the SD+PM setting would be the most ideal as it provides the most novel designs. In contrast, if a designer wants to explore alternative designs in the later stages of the design process, producing feasible images may be a priority over highly novel ones. Therefore, CIP(0.51), CIP(0.67), or CIP(0.83) settings may be suitable for this scenario. 

Based on the results, we offer a few points of caution when selecting the appropriate setting for image generation. First, selecting a CAD image prompt weight higher than 0.83 is likely disadvantageous. Past 0.83, the perceived feasibility of the generated images begins to drop, perhaps because the T2I model’s ability to balance adherence to the CAD image with the text prompt is compromised. Secondly, the decrease in image novelty seems to plateau at the CAD image prompt weight of 0.67. This means that past 0.67, about 0.83 may always be the ideal setting as it provides the most visibly feasible designs within that range. It is important to note that the specific values of the image prompt weight mentioned in these suggestions may be model specific. Therefore, if a T2I model other than Stable Diffusion 2.1 is being used, the appropriate values for that model must be investigated. Lastly, if no CAD image prompting is used, the SD+PM setting may always be better than the SD setting in terms of feasibility and novelty. SD+PM produces designs that are significantly more feasible yet similar in novelty.

\subsection{Limitations and Future Work}
This section describes the limitations of this work and their corresponding future research directions. The first limitation is that the performance of the method proposed in this paper may be dependent on the dataset. The CAD images that guide image generation are selected from a dataset of CAD images. This means that the quality and the prompt conformity of the CAD images are determined by the dataset. This issue may be resolved by exploring alternative methods to produce the CAD images for prompting, such as designers manually creating them using CAD software.

In addition, there is an inherent trade-off between the T2I model's adherence to the text and image prompts. With a higher image prompting weight, the T2I model is less faithful to the text prompt requirements, and vice versa. For example, in \ref{tab:2}, the generated bike images in the higher image prompt weight settings tend to exclude some required elements from the text prompts like wings and sofa chair. Investigating this tradeoff and its impact on the designers and their design process can add more insights into how our CAD image prompting method can be effectively utilized in engineering design. 

Furthermore, we present one case study in this paper that serves as an initial exploration and proof-of-concept of our proposed method. Further validation of the method is a valuable direction for future research. 

Finally, other potential future works include assessing the perceived feasibility and novelty of the CAD images themselves as a benchmark for comparison with the other generation settings and empirically investigating the difficulties engineers and designers currently face using T2I models and identifying how our CAD image prompting may help those challenges. 

\section{Conclusion}
In engineering design, T2I models are difficult to be utilized as more than an inspirational tool during concept generation because they produce infeasible designs. This paper proposes a CAD image prompting method that aims to improve the perceived feasibility of designs generated by T2I models. The effectiveness of this method was explored through a case study with a bike design task, which showed that CAD image prompting successfully leads to more visibly feasible designs. A feasibility-novelty tradeoff was also observed, suggesting that caution and strategy is needed when selecting the weight of the CAD image prompt. When utilized effectively, our method has the potential to expand the range of applicability of T2I models in the engineering design process. Moreover, our CAD image prompting method possesses unique advantages due to its use of CAD images. Particularly, when combined with image-to-CAD methods, it can lead to the development of a computational design tool that supports multimodal design, enabling designers to easily switch back and forth between concept generation and design refinement.

%%%%% Acknowledgments %%%%%%%%%%%%%%%%%%%%%%%%%%%

\section*{Acknowledgments}
Prof. Ahmed and Prof. Lykourentzou extend their gratitude to the MIT MISTI Netherlands Program for their funding support.

%%%  REFERENCES  %%%%%%%%%%%%%%%%%%%%%%%%%%%%%%%%

\nocite{*}%% <=== Delete this line unless you want to typeset the entire contents of your .bib file!

\bibliographystyle{asmeconf}  %% .bst file following ASME conference format. Do not change.
\bibliography{bib}%% <=== change this to name of your bib file

%%%%%%%%%%%%%%%%%%%%%%%%%%%%%%%%%%%%%%%%%%%%%%%%%%%%%%%%%%%%%%%%%%%%%%%%%%%%%%%%%%%%%%%

\end{document}